\documentclass[10pt,twocolumn,letterpaper]{article}

\usepackage{cvpr}              
\usepackage[accsupp]{axessibility}

\usepackage{graphicx}
\usepackage{amsmath}
\usepackage{amssymb}
\usepackage{booktabs}
\usepackage{multirow}
\usepackage{mathrsfs}
\usepackage{soul}
\usepackage[normalem]{ulem}

\usepackage[pagebackref,breaklinks,colorlinks]{hyperref}
\usepackage{mathrsfs}

\usepackage[capitalize]{cleveref}
\crefname{section}{Sec.}{Secs.}
\Crefname{section}{Section}{Sections}
\Crefname{table}{Table}{Tables}
\crefname{table}{Tab.}{Tabs.}

\newcommand{\ho}{{hand-object~}}

\newcommand{\bx}{{\mathbf{x}}}

\def\cvprPaperID{8445} 
\def\confName{CVPR}
\def\confYear{2023}

\begin{document}

\title{Overcoming the Trade-off Between Accuracy and Plausibility \\ in 3D Hand Shape Reconstruction}
\author{Ziwei Yu$^1$  Chen Li$^1$  Linlin Yang$^1$ Xiaoxu Zheng$^2$ Michael Bi Mi$^2$  Gim Hee Lee$^1$ Angela Yao$^1$\\
$^{1}$National University of Singapore   $\quad$        $^{2}$Huawei International Pte Ltd, Singapore\\
{\tt\small \{yuziwei, lichen\}@u.nus.edu}, {\tt\small \{mu4yang, zhengxiaoxu66\}@gmail.com}, {\tt\small michaelbimi@yahoo.com} \\{\tt\small \{gimhee.lee, ayao\}@comp.nus.edu.sg}}
\maketitle

\begin{abstract}
Direct mesh fitting for 3D hand shape reconstruction is highly accurate. However, the reconstructed meshes are prone to artifacts and  do not appear as plausible hand shapes.  Conversely, 
parametric models like MANO ensure plausible hand shapes but are not as accurate as the non-parametric methods.  In this work, we introduce a novel weakly-supervised hand shape estimation framework that integrates non-parametric mesh fitting with MANO model in an end-to-end fashion.  
Our joint model overcomes the tradeoff in accuracy and plausibility to yield well-aligned and high-quality 3D meshes, especially in challenging two-hand and hand-object interaction scenarios.
\end{abstract}

\section{Introduction}
\begin{figure}[!ht]
	\centering{
	\includegraphics[width=1.0\linewidth]{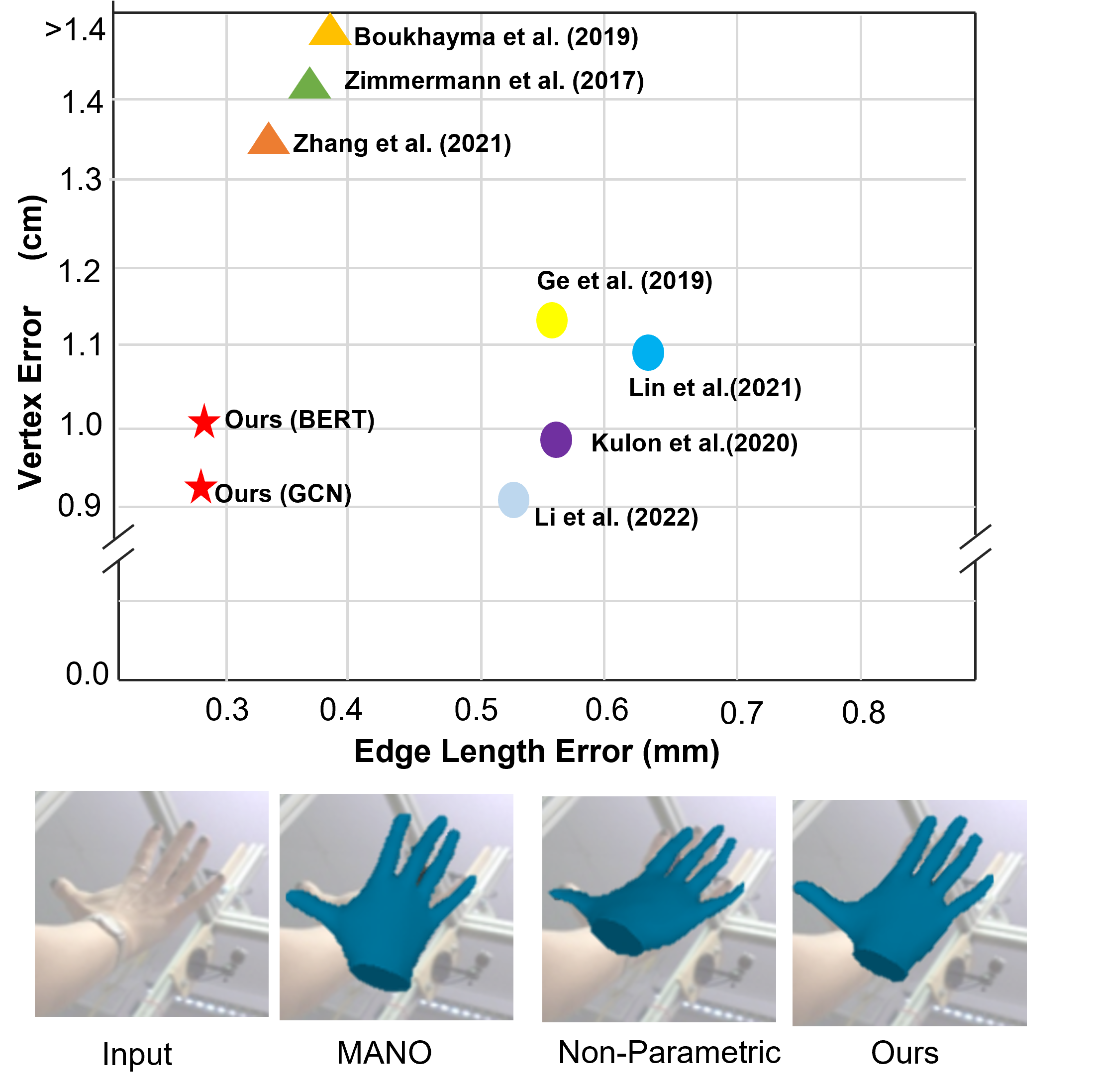}
	\caption{ The vertex error vs. edge length error indicates that existing methods trade-off alignment accuracy with plausibility.  MANO-based methods (triangles) vs. non-parametric model-based methods (circles) have a trade-off between vertex error and edge length error; our combined method (stars) can overcome this trade-off to yield well-aligned and plausible meshes.  Plot of results from InterHand2.6M~\cite{Moon_2020_ECCV_InterHand2.6M}; visualization from FreiHand~\cite{zimmermann2019freihand}. 
    }
	\label{fig:teaser}}
\end{figure}

State-of-the-art monocular RGB-based 3D hand reconstruction methods~\cite{bib:MobRecon,ge2019handshapepose,lin2021end-to-end,Li2022intaghand,Kulon_2020_CVPR,park2022handoccnet} focus on recovering highly accurate 3D hand meshes. As accuracy is measured by an average joint or vertex position error, recovered hand meshes may be well-aligned in 3D space but still be physically implausible.  The 3D mesh surface may have irregular protrusions or collapsed regions~(see Fig. \ref{fig:teaser}), especially around the fingers. The meshes may also suffer from incorrect contacts or penetrations when there are hand-object or two-hand interactions.  Yet methods that prioritize physical plausibility, especially in interaction settings~\cite{ge2019handshapepose,li2021hybrik,hasson19_obman,Cao2021,chen2022alignsdf,Li2022intaghand}, are significantly less accurate in 3D alignment.  In summary, the current body of work predominantly favours either 3D alignment accuracy or physical plausibility, but cannot achieve both.

A closer examination reveals that the trade-off between 3D alignment and plausibility is also split methodology-wise.  Methods that use the MANO model~\cite{MANO:SIGGRAPHASIA:2017} produce plausible hand poses and hand shapes~\cite{zimmermann2019freihand,boukhayma20193d,zhang2019end,chen2021model,Zhang2021twohand} due to the statistical parameterization of MANO.  However, it is challenging to directly regress these parameters, since the mapping from an image to the MANO parameter space is highly non-linear.  As a consequence, MANO-based methods lag in 3D alignment accuracy compared to non-parametric methods.

Non-parametric methods~\cite{bib:MobRecon,ge2019handshapepose,lin2021end-to-end,Li2022intaghand,Kulon_2020_CVPR,park2022handoccnet,gong2019spiralnet++,li2021coarse} directly fit a 3D mesh to image observations.  Direct mesh fitting is accurate but is prone to surface artifacts. In scenarios with hand-object or hand-hand interactions, mesh penetrations cannot be resolved meaningfully even with regularizers such as contact losses~\cite{hasson19_obman} due to the unconstrained optimization.  
Attention mechanism~\cite{tse2022collaborative,Li2022intaghand} can mitigate some penetrations and artifacts, but the inherent problem remains. As such, the favoured approaches for hand-object and hand-hand interactions are still driven by MANO models~\cite{hasson19_obman,hasson20_handobjectconsist,chen2022alignsdf,Zhang2021twohand,Cao2021}.

In this work, we aim to recover high-quality hand meshes that are accurately aligned \emph{and} 
have minimal artifacts and penetrations. 
To avoid a trade-off, we leverage direct mesh fitting for alignment accuracy and guidance from MANO for plausibility. Combine the non-parametric and parametric models is straightforward in terms of motivation.  However, merging the two is non-trivial because it requires a mapping from non-parametric mesh vertices to parametric model parameters.  This mapping, analogous to the mapping from an RGB image, is highly non-linear and difficult to achieve directly~\cite{kolotouros2019convolutional}. 

One of our key contributions in this work is a method to accurately map non-parametric hand mesh vertices to MANO joint parameters $\theta$. To do so, we perform a two-step mapping, from mesh-vertices to the joint coordinates, and joints coordinates to $\theta$. In the literature, the common practice for the former is to leverage the $\mathcal{J}$ matrix in MANO and linearly regress the joints from the mesh~\cite{Kulon_2020_CVPR,Li2022intaghand,li2021hybrik}.  Yet the $\mathcal{J}$ matrix was designed to only map MANO-derived meshes to joints in a rest pose (see Eq. 10 in~\cite{SMPL:2015}). As we show in our experiments, applying $\mathcal{J}$ to non-rest poses introduces a gap of around 2 mm. Furthermore, we postulate that there is a domain gap between the estimated non-parametric meshes and MANO-derived meshes, even if both meshes have the same topology.  To close this gap -- we propose a VAE correction module, to be applied after the linear regression with 
$\mathcal{J}$.
To map the recovered joints from the mesh to $\theta$, we use a twist-swing decomposition and analytically compute the $\theta$. It has been shown previously in ~\cite{li2021hybrik} that decomposing joint rotations into twist-swing rotations~\cite{baerlocher2001parametrization} can simplify the estimation of human SMPL~\cite{SMPL:2015} model pose parameters. 
Inspired by \cite{li2021hybrik}, we also leverage the decomposition and further verify that the twist angle has minimal impact on the hand.  

\begin{figure*}[!htb]
	\centering{
	\includegraphics[width=1.0\linewidth]{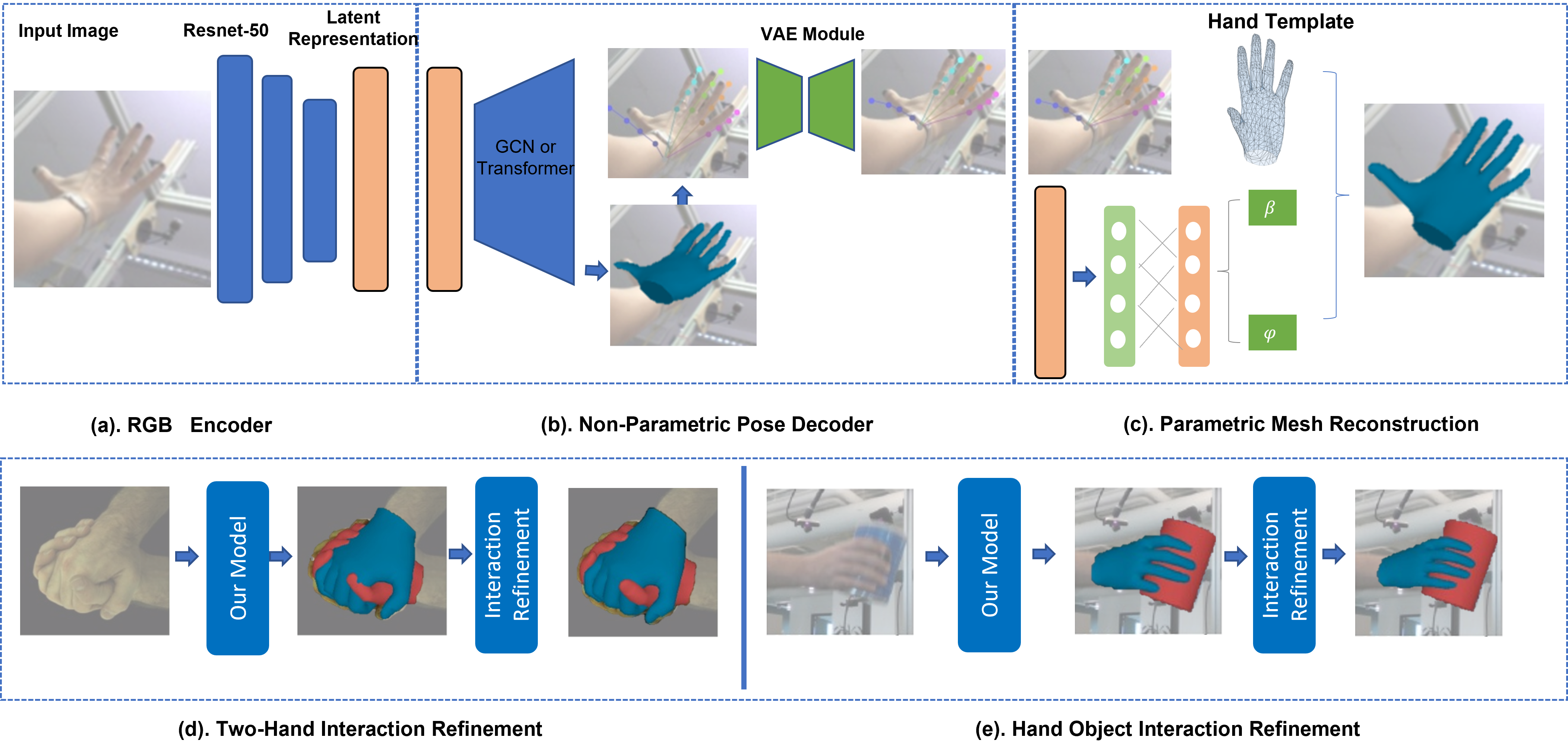}
	\caption{Overview of our pipeline integrating a non-parametric model and a MANO model. Our proposed framework has an RGB image encoder (a), a non-parametric pose decoder (b), and a parametric mesh reconstruction (c). We also show our proposed pipeline success when used in two-hand interaction refinement in (d) and hand-object interaction refinement in (e).}
	\label{fig:framework}}
\end{figure*}

Note that obtaining ground truth labels for hand mesh vertices is non-trivial.   
Our framework lends itself well for weak-supervision. Since the estimated 3D mesh from the non-parametric decoder is regressed into 3D joints, it can also be supervised with 3D joints as weak labels (see Fig \ref{fig:framework}).  
At the same time, the parametric mesh estimated from these joints can be used as a pseudo-label for learning the non-parametric mesh vertices.  Such a procedure distills the knowledge from the parametric model and is effective without ground truth mesh annotations.  We name our method \textbf{WSIM 3D Hand}, in reference to \textbf{W}eakly-supervised \textbf{S}elf-distillation \textbf{I}ntegration \textbf{M}odel for \textbf{3D} \textbf{hand} shape reconstruction.  Our contributions include:

\begin{itemize}
\item A novel framework that integrates a parametric and a non-parametric mesh model for accurate \emph{and} plausible 3D hand reconstruction.

\item A VAE correction module that closes the overlooked gap between non-parametric and MANO 3D poses; 

\item A weakly-supervised pipeline, competitive to a fully-supervised counterpart, using only 3D joint labels to learn 3D meshes.  

\item Significant improvements over state-of-the-art on hand-object or two-hand interaction benchmark datasets, especially in hand-object interaction on DexYCB.
\end{itemize}

\section{Related Work}
\paragraph{Parametric Methods.}
Previous works~\cite{zhang2019end,boukhayma20193d,chen2021model,yu2021local,hasson19_obman,Cao2021,yang2021cpf,hasson20_handobjectconsist,zhang2020phosa,Zhang2021twohand} in 3D hand shape reconstruction often used the MANO model~\cite{MANO:SIGGRAPHASIA:2017}  to estimated 3D hand meshes. Boukhayma et al.~\cite{boukhayma20193d} first use a deep neural network to regress the MANO parameters in single-hand reconstruction. However, directly estimating MANO parameters accurately is challenging, as they sit in an abstract PCA space. Moreover, previous MANO model-based methods ignore the spatial information that limited their reconstruction accuracy~\cite{chen2021i2uv,ge2019handshapepose}. This work addresses the above drawbacks of the MANO model by integrating a non-parametric model.

\paragraph{Non-parametric Methods}
Non-parametric 3D hand shape methods~\cite{ge2019handshapepose,Kulon_2020_CVPR,bib:MobRecon,Li2022intaghand,tse2022collaborative,lin2021end-to-end,park2022handoccnet,hampali2022keypoint,liu2022spatial,yu2022UV}
directly fit the mesh vertices either with graph convolutional networks~\cite{cnn_graph,gong2019spiralnet++} or transformers~\cite{vaswani2017attention}. Initially, spectral graph neural networks were used~\cite{cnn_graph} but as they are not able to leverage deeper neighbourhood nodes' information, spatial graphs with spiral convolutions~\cite{Kulon_2020_CVPR} were proposed instead.  Subsequently, works applied mesh transformers~\cite{lin2021end-to-end} and other attention mechanisms~\cite{park2022handoccnet,hampali2022keypoint,liu2022spatial} to facilitate interaction modelling. The estimated 3D pose and shapes of non-parametric methods are highly accurate; however, their many degrees of freedom also yield implausible 3D shapes with artifacts. This work integrates non-parametric methods with a MANO model to achieve both alignment accuracy and plausibility.

\paragraph{Hand-Object and Two-Hand Interactions}
Hand interactions add challenge to 3D shape reconstruction due to the additional occlusion from the interacting object or hand and possibility of surface collisions.  For hand-object interaction, previous works~\cite{hasson19_obman,Cao2021,yang2021cpf,hasson20_handobjectconsist,zhang2020phosa} leverage the MANO model to ensure plausible hand shapes during the interaction modelling.  Similarly, for two-hand interactions~\cite{Moon_2020_ECCV_InterHand2.6M,Zhang2021twohand}, MANO has been applied to the left and right hand individually to simplify the two-hand reconstruction into two single hand parameters estimation pipeline.
By using MANO in the interaction setting, these above works are able to estimate plausible 3D hand shapes, though the alignment accuracy generally lags compared to non-parametric methods~\cite{Li2022intaghand,tse2022collaborative,bib:MobRecon} that are less plausible.
Different from the above methods, we first use the non-parametric model to learn the 3D joints and then convert these joints into accurate MANO parameters in the interaction setting. Therefore, our work overcomes the tradeoff between plausibility and accuracy.

\section{Method}
Fig.~\ref{fig:framework} shows an overview of our method.  It has three components: the RGB encoder network, a non-parametric pose decoder~(Sec.~\ref{subsec:GCN}) and a parametric mesh reconstruction~(Sec.~\ref{subsec:Parametric}) based on the MANO model~(Sec.~\ref{subsec:MANO}). The overall framework can be learned in a weakly-supervised manner with self-distillation~(Sec.~\ref{subsec:dis}).  For hand-object and hand-hand interactions, we add an interaction refinement module to reduce the penetration~(Sec.~\ref{subsec:inter}). 

\subsection{Preliminaries}\label{subsec:MANO}
MANO~\cite{MANO:SIGGRAPHASIA:2017} is a statistical 3D hand pose and shape model. 
 It maps pose parameters $\theta \in R^{16 \times 3}$ and shape parameters $\beta \in R^{10}$ to a 3D hand mesh  with model $M(\beta, \theta)$:
\begin{equation}
\begin{aligned}
      T(\beta, \theta) = \hat{T} + B_S(\beta) +  B_P(\theta),\\ 
    M (\beta, \theta) = W(T(\beta, \theta), J(\beta), \theta, \mathcal{W}).\label{eq:mano}
\end{aligned}
\end{equation}
\noindent The hand template $T(\beta, \theta) \in R^{778 \times 3}$, also called the T-template, is obtained by deforming a mean mesh $\hat{T} \in R^{778 \times 3}$ with shape and pose corrective blend shapes, $B_S(\beta)$ and $B_P(\theta)$. The hand template can be converted into the reference `rest' T-pose: $J(\beta) \in R^{16 \times 3} $ with a linear regression using $J(\beta) = \mathcal{J}\times T(\beta, \theta)$, where the $\mathcal{J} \in  R^{16 \times 778}$ matrix stores the regression weights. The blend function $W$ returns the mesh and joint from the T-template, T-pose and $\theta$ parameters, where $\mathcal{W} \in R^{778 \times 16}$ is a linear blend skinning matrix. With this MANO mesh model, we can easily reconstruct a hand mesh by using specific shape parameters $\beta$ and pose parameters $\theta$.   

An often overlooked point for MANO is that the regression weight in the matrix $\mathcal{J}$ is only designed for the rest or T-pose $J(\beta)$.  It should not be applied to regress the pose from arbitrary meshes $M(\beta, \theta)$, even though this is done by several existing works~\cite{Kulon_2020_CVPR,Li2022intaghand,li2021hybrik}. In fact, these works also apply it to non-MANO derived meshes. Our experimental results show a nearly 2 mm gap when $\mathcal{J}$ is used in this way (see Supplementary Section~2).

\textbf{Twist-Swing Decomposition:} Directly regressing the pose parameters $\theta \in R^{16 \times 3}$ is challenging due to the MANO model uses the kinematic chain scheme to adjust joint rotation, that the accumulated error raises the learning challenge~\cite{ge2019handshapepose, chen2021i2uv}. Furthermore,  pose parameters $\theta \in R^{16 \times 3}$ represent the rotation matrix of the hand joints $J$ in SO3 space. However, directly regressing these rotation parameters is an ambiguous problem due to these rotations being non-continuous~\cite{Zhou2018on}. The previous work~\cite{baerlocher2001parametrization} showed that twist-swing decomposition is an effective way for a ball-and-socket joint system to reduce the learning difficulty. Therefore, instead of directly regressing the pose parameters $\theta$, we leverage the twist-swing decomposition and combine the hand joint locations to recover more accurate pose parameters.
 
\subsection{Non-parametric Pose Decoder}\label{subsec:GCN}
Like previous works~\cite{ge2019handshapepose, Kulon_2020_CVPR, Li2022intaghand}, we adopt a ResNet50~\cite{He2015} as the backbone and encode the input image $I\in R^{256 \times 256}$ into a latent feature $z \in R^{1000}$. The latent feature $z$ is passed into a pose decoder to obtain the hand joints $J_{\text{pred}} \in R^{21 \times 3}$.
\begin{equation}
    \hat{J} = \mathcal{J}\times M_{\text{np}}(V,F), \;\; \text{where}\;\;  M_{\text{np}}(V, F) = f(z)\label{eq:non}.
\end{equation}
Specifically, a hand mesh $M_{\text{np}} = (V, F)$ with vertices $V \in R^{778 \times 3}$ and faces $F \in R^{1538 \time 3}$ is estimated from the latent feature $z$ with a non-parametric model $f$.  The model $f$ can be any off-the-shelf non-parametric method, \eg based on GCNs~\cite{ge2019handshapepose, Kulon_2020_CVPR, Li2022intaghand}, or transformers~\cite{lin2021end-to-end}. The mesh $M_{\text{np}}$ must have the same topology as the MANO model. Following~\cite{Kulon_2020_CVPR,Li2022intaghand,li2021hybrik}, we use the MANO $\mathcal{J}$ matrix to regress hand joints $\hat{J} \in R^{21 \times 3}$ from the mesh, even though it is not intended as such and will introduce a performance gap. To bridge this unwanted gap, 
we add a VAE module to refine the $\hat{J} \in R^{21 \times 3}$ to  $\widetilde{J} \in R^{21 \times 3}$. 

Our proposed VAE module consists of a full connection layer as an encoder and symmetric full connection layers as the decoder.
Then, the pose decoder can be learned with the following loss function:
\begin{equation}
L_{\text{joint}} =  ||\hat{J} - J || +  ||\widetilde{J} - J || + ||\hat{J} - \widetilde{J} || + \lambda_1 * L_{KL} 
\end{equation}
where $J$ is the ground truth joints and $L_{KL} = KL(q(z|\hat{J})||p)$ is the standard Kullback-Leibler divergence loss used in VAE models, where $z$ represents the latent variables encoded from input $\hat{J}$. The term $p = \mathcal{N}(0, E)$ denotes a Gaussian prior where $E$ is an identity matrix. 

\subsection{Parametric Mesh Reconstruction}\label{subsec:Parametric}
Directly regressing accurate MANO parameters is highly challenging due to their abstract nature. Twist-swing decomposition is an alternative way to learn these parameters as we already discussed.
Further, Li et al.~\cite{li2021hybrik} introduced the use of twist-swing decomposition to overcome the challenge of the analogous SMPL~\cite{SMPL:2015} model for human body pose and shape estimation. 
We follow their setting to infer the MANO pose parameters by using the hand joints and twist parameters, like $\theta = f(\varphi, \widetilde{J})$,
where $\varphi \in R^{16 \times 3}$ is the hand joint twist rotation matrix. 

An interesting finding from our experiments is that there is no difference when using the joint or vertex as supervision for the MANO parameter learning. See Sec.~1 in the Supplementary Material for more details.

These results further demonstrate the possibility of employing joints as weak labels for the learning of MANO mesh.
Therefore, we can utilize the estimated joint $J_{\text{pred}}$ and estimated twist rotation matrix $\varphi$ to predict the MANO pose parameter $\theta$. After that, the estimated $\beta$ and $\theta$ are fed into the MANO model to obtain a final well-aligned and plausible hand mesh. For the parametric mesh reconstruction, we use a regularized L1 loss with respect to the ground truth:
\begin{equation}
\begin{aligned}
     L_{\text{shape}} =  ||\beta_{\text{pred}} - \beta_{\text{gt}} || + ||\beta_{\text{pred}}||,\\
    L_{tw} =  ||\varphi_{\text{pred}} - \varphi_{\text{gt}}|| + ||\varphi_{\text{pred}} ||,
\end{aligned}
\end{equation}
where $\beta_{\text{gt}}$ and $\varphi_{\text{gt}}$ are the ground truth shape parameters and twist parameters, respectively.

\subsection{Interaction Refinement}\label{subsec:inter}
In our interaction setting, similar to previous works~\cite{rong2021monocular, hasson19_obman,Cao2021,chen2022alignsdf}, we first estimate the two hands, or the hand and the object, individually, then we add a refinement module.
To reduce the penetration for the two-hand interaction, same as~\cite{rong2021monocular, Cao2021, chen2022alignsdf}, we use a Signed Distance Field (SDF) from one hand mesh to check whether the vertex on another hand or object is inside this hand mesh. The SDF is obtained by voxelizing the left and the right hand meshes (two-hand interaction) or hand and object meshes (hand-object interaction) to a $32\times32\times32$ 3D grid. Then, the modified SDF function $\phi$ for this hand mesh can be written as follows:
\begin{equation}\label{eq:SDF}
      \phi(x,y,z) = - min(\text{SDF}(c_x, c_y, c_z),0).
\end{equation}
For each cell in the 3D grid $c = (c_x,c_y,c_z)$, the $\phi(x,y,z)$ takes positive values if the cell is inside the hand mesh, and zero if outside. 

For two-hand interaction, the loss is calculated as follows: 
\begin{equation}\label{eq:interaction}
      L_{\text{pene}} =\frac{1}{|V_{\text{in}}^r|}\sum_{\bx_V \in V_{\text{in}}^r}\text{dist}(\text{v},V_r) + \frac{1}{|V_{\text{in}}^l|}\sum_{\bx_V \in V_{\text{in}}^l} \text{dist}(\text{v},V_l),
\end{equation}
where $V_{\text{in}}^r$ refers to vertices from the left hand which has penetrated into the right hand, and vice versa for $V_{\text{in}}^l$.
The dist($\cdot$) represents the minimal distance between the inside vertex and the hand surface. As for the hand-object interaction, we use the same penetration loss in our hand-object interaction refinement following ~\cite{Cao2021},; please refer to~\cite{Cao2021} for more details. 

\subsection{Weak Label \& Self-Distillation}\label{subsec:dis}
Obtaining ground truth 3D mesh vertices is non-trivial, hence we propose a weakly-supervised approach that uses the 3D joints instead. Recently, self-distillation has become popular for unsupervised pose estimation~\cite{li2021synthetic,liu20223d,ren2021spatial}.
We follow a similar approach, under the assumption that the parametric reconstruction, which is strongly supervised with ground truth hand joints and MANO parameters, yields more accurate 3D meshes which can be distilled to the non-parametric branch. We use an L1 loss for self-distillation:
\begin{equation}
    L_{\text{vert}} =  ||V_{refine} - V_{non}||,
\end{equation}
where the $V_{refine} \in R^{778 \times 3}$ and $V_{non} \in R^{778 \times 3}$ are the hand vertices from our parametric mesh reconstruction module and non-parametric model, respectively.
The overall loss function of our proposed pipeline in single-hand shape reconstruction is formulated as: 
\begin{equation}
    L_{\text{total}} = \lambda_2 L_{\text{shape}} + \lambda_3 L_{tw} + \lambda_4 L_{\text{joint}} + \lambda_5 L_{\text{vert}}.
\end{equation} 

In two-hand interaction and hand-object interaction refinement, we follow~\cite{hasson19_obman} and use the above loss function to train the whole pipeline. After that, we use a small learning rate $1e{-6}$ and $L_{\text{pene}}$ to do the interaction refinement; the interaction loss function can be written as follows:
\begin{equation}
    L_{\text{inter}} = L_{\text{total}} + \lambda_6 L_{\text{pene}}.
\end{equation}

\section{Experimental Results}
\subsection{Implementation Details} {Our network consists of three independent modules: image feature extraction, a non-parametric pose decoder and a parametric mesh reconstruction. To ensure a fair comparison, all of our experiments use the same pretrained ResNet 50~\cite{He2015} as a backbone to extract the input image feature. For the non-parametric pose decoder part, we consider three state-of-the-art structures, \ie, a Graph Convolution Network, a Mesh Transformer Network and a MANO layer network, which  are the same as in~\cite{bib:MobRecon},~\cite{lin2021end-to-end},~\cite{hasson19_obman}, respectively. The parametric mesh reconstruction module consists of one VAE network and one differentiable layer to calculate the rotation matrix based on given joints. For the shape parameter and twist parameter prediction, two fully-connected networks are used.  The above hyper-parameters are set empirically to $\lambda_1 = 0.001, \lambda_2 = 10, \lambda_3 = 10, \lambda_4=100, \lambda_5=100, \lambda_6= 10$. The dimension of VAE latent space is 128.}

\subsection{Training Details}
{The Adam optimizer is applied to train all networks over 200 epochs with a batch size of 64. We start with an initial learning rate of $10^{-4}$ for all training settings and lower it by a factor of 10 at the 50th, 100th, and 150th epochs. After JointVAE is trained, we apply this pretrained JointVAE in our pipeline. We set all of the hypermeters $\lambda$ empirically. In two-hand or hand-object interaction refinement, same as~\cite{hasson19_obman}, after training the whole pipeline, we use a learning rate of $1e-6$ and a physical contact loss to refine the two-hand and object interaction after the 10th epoch. }

\begin{table*}[htb]
\centering
\scalebox{1.0}
{
\begin{tabular}{p{3.0cm}|p{0.7cm}p{0.7cm}p{0.7cm}p{0.7cm}|p{0.7cm}p{0.7cm}p{0.7cm}p{0.7cm}|p{0.7cm}p{0.7cm}p{0.7cm}p{0.7cm}} 
\hline
\multicolumn{1}{l|}{Dataset} &\multicolumn{4}{c|}{FreiHAND}  &\multicolumn{4}{c|}{InterHand} &\multicolumn{4}{c}{DexYCB} \\ 
\cline{1-13} 
Method &\scriptsize{MPJPE} &\scriptsize{MPVPE} &\scriptsize{Edge} &\scriptsize{Norm}  &\scriptsize{MPJPE} &\scriptsize{MPVPE}  &\scriptsize{Edge} &\scriptsize{Norm}   &\scriptsize{MPJPE} &\scriptsize{MPVPE}  &\scriptsize{Edge} &\scriptsize{Norm}
\\\hline

Zhang et al.~\cite{Zhang2021twohand}&- &- &- &- &13.48 &13.95 &0.31 & 0.15  &- &- &- &-\\
Hasson et al.~\cite{hasson19_obman}&13.3 &13.3 &0.68 &0.17 &14.21 &- &- &-  &- &- &- &-\\
Moon et al.~\cite{moon2020interhand2}&- &- &- &- &14.21 &- &- &-  &- &- &- &-\\
Rong et al.~\cite{rong2021monocular}&- &- &- &- &17.12 &- &- &-  &- &- &- &-\\
Li et al.~\cite{Li2022intaghand}&- &- &- &- &\textbf{8.79} &\textbf{9.03} &0.51 &0.12  &- &- &- &-\\
Li et al.~\cite{Li2022intaghand} Baseline &- &- &- &- &9.97 &10.63 &0.51 &0.12  &- &- &- &-\\
Boukhayma et al.~\cite{boukhayma20193d} &13.08 &13.40 &0.64 &0.17 &16.93 &17.98 &0.31 & 0.15  &12.88 &12.98 &.
48&0.15\\
MANO CNN~\cite{zimmermann2019freihand}&8.69 &8.83 &0.54 &0.16 &13.87 &14.27 &0.32 & 0.16  &10.68 &11.61 &0.30 &0.14\\
GCN-Vert~\cite{Kulon_2020_CVPR}&7.77  &7.43 &0.94 &0.20  &9.95 &10.23  &0.56 &0.13  &\underline{8.93} &\underline{9.39} &0.51 &0.15\\ 
Transformer~\cite{lin2021end-to-end}  &7.57 &8.05 &0.81 &0.16  &10.89 &10.83 &0.68 &0.15  &9.51 &10.48 &0.65 &0.15\\ \hline
MANO-Joint~\cite{zimmermann2019freihand}&8.84 &9.10 &0.55 &0.17  &13.98 &14.35 &0.32 & 0.17  &15.44 &16.15 &0.42 &0.22\\
GCN-Joint~\cite{Kulon_2020_CVPR} &14.87  &18.43 &3.81 &0.34  &11.49 &19.20 &5.10 &0.43  &10.07 &15.12 &3.49 &0.31\\  \hline
Ours (GCN)&\underline{7.42} &\underline{7.43} &\underline{0.51} &\underline{0.15} &\underline{9.68} &\underline{9.89} &\textbf{0.27} &\textbf{0.12}  &\textbf{8.92} &\textbf{9.12} &\textbf{0.25} &\textbf{0.12}\\ 
Ours (Trans)  &\textbf{7.28} &\textbf{7.33} &\textbf{0.49} &\textbf{0.15} &10.08 &10.06 &\underline{0.29} &\underline{0.13}  &9.13 &9.67 &\underline{0.28} &\underline{0.14}\\ \hline
\end{tabular}}
\caption{Comparisons with state-of-the-art methods on the FreiHAND, InterHand and DexYCB test sets. \textbf{Best} and \underline{second-best} scores. 
Ours (GCN) and Ours (Trans) achieve the best or second-best holistic performance across all comparisons  
.}
\label{tab:quantitative_table1}
\end{table*}  

\subsection{Datasets and Evaluation Metrics}
\paragraph{Datasets.} {Our method is evaluated on three types of RGB-based \ho benchmarks, \ie, the One Single hand shape reconstruction dataset on FreiHAND~\cite{zimmermann2019freihand}, to evaluate the hand shape of our integrated model in single-hand tasks. {FreiHAND} is a challenging multi-view RGB dataset of \ho interactions that contains 37k samples of hands manipulating objects. The second one is two-hand shape reconstruction dataset on Interhand2.6M~\cite{Moon_2020_ECCV_InterHand2.6M} to evaluate our pipeline for mesh reconstruction on a two-hand interaction dataset. Interhand2.6M is a two-hand interaction dataset. We use a dataset setting similar to that of \cite{Li2022intaghand}, which consists of 366K training samples and 261K testing samples. The last one is  hand-object interaction dataset, DexYCB~\cite{chao:cvpr2021}, the latest large-scale RGB-based \ho dataset. It contains 582k samples of hands grasping 20 YCB objects, and is used here  to evaluate our proposed model on hand-object interaction.  We evaluate our method using the official ``S0'' split.  The \ho images in this dataset contain 10 objects modeled from YCB objects~\cite{xiang2017posecnn}. We compare our methods with the state-of-the-art on both of these versions and report our results through their leaderboards. All input images are cropped and resized to $256 \times 256$ based on their 2D projection.}

\paragraph{Metrics.} {To evaluate the accuracy of our predicted 3D hand pose and surface, we use the mean-per-joint-position-error (MPJPE) for 3D joints and the mean-per-vertex-position-error (MPVPE) for mesh vertices. In addition, unlike previous works that only focus on vertex accuracy for mesh evaluation, we introduce the edge error distance (mm) and normal error~\cite{ge2019handshapepose} as extra evaluation metrics to evaluate the plausibility of hand mesh. 
Meanwhile, in two-hand interaction, we calculate the penetration depth (PD) of each hand vertices penetrating into the other hand in the 3D grid ($32\time32\times32$) above. There are two types of penetration depth, \ie, Average Penetration Depth (A-PD) and Maximum Penetration Depth (M-PD). As for hand-object interaction, the penetration distance calculation is the same as for two-hand interaction. }

\begin{figure*}[!htb]
	\centering{
	\includegraphics[width=1.0\linewidth]{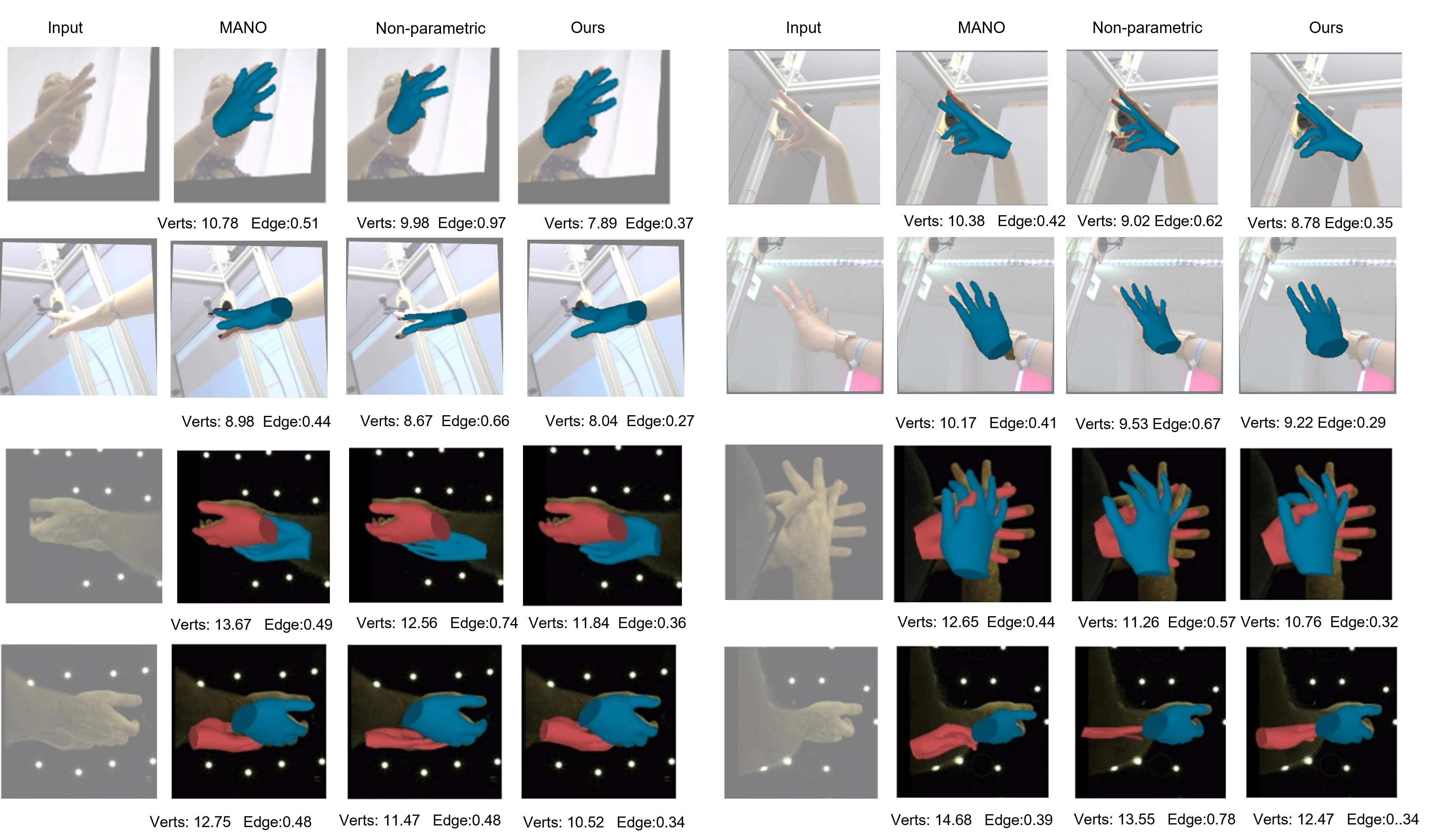}
	\caption{Hand shape reconstruction results. For each quartet, from left to right columns correspond to RGB input, MANO based method: MANO CNN~\cite{zimmermann2019freihand}, non-parametric model based method: GCN-vert~\cite{Kulon_2020_CVPR} and our method in camera view. Besides, vertex error and edge length error are also reported for quantitative evaluation.  Low vertex error and edge length error indicate well-aligned and plausible hand meshes.}
	\label{fig:qualitative1}}
\end{figure*}

\begin{figure}[!htb]
	\centering{
	\includegraphics[width=1.0\linewidth]{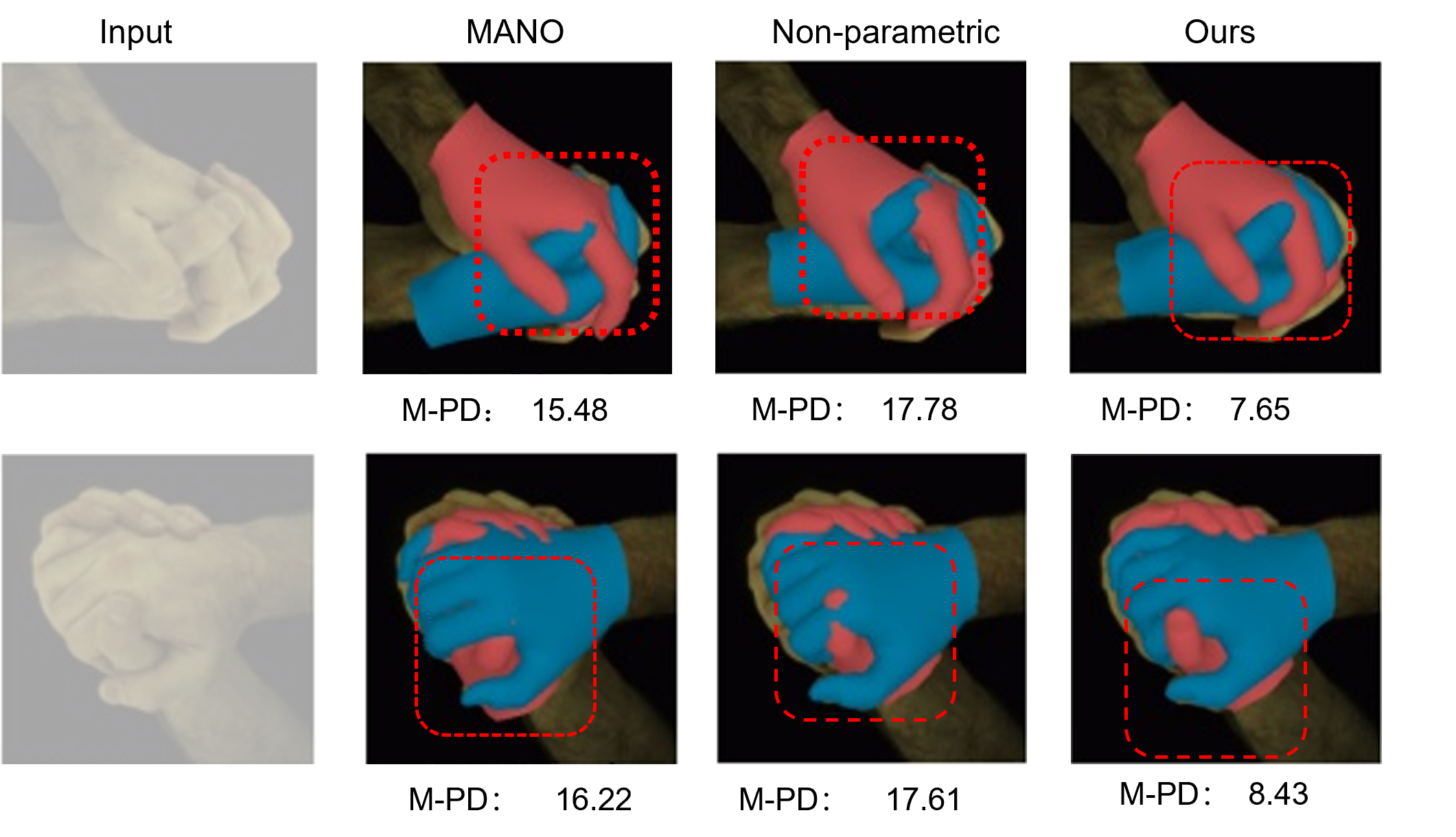}
	\caption{Two-hand reconstruction results.  
 For each quartet, left to right  columns correspond to input RGB images, MANO CNN~\cite{zimmermann2019freihand}, non-parametric model based method~\cite{Li2022intaghand} and our mesh. The red box highlights the penetration region and we report the max penetration depth values for quantitative evaluation. Our proposed method yields more plausible two-hand interactions.}
	\label{fig:qualitative2}}
\end{figure}

\begin{figure}[!htb]
	\centering{
	\includegraphics[width=1.0\linewidth]{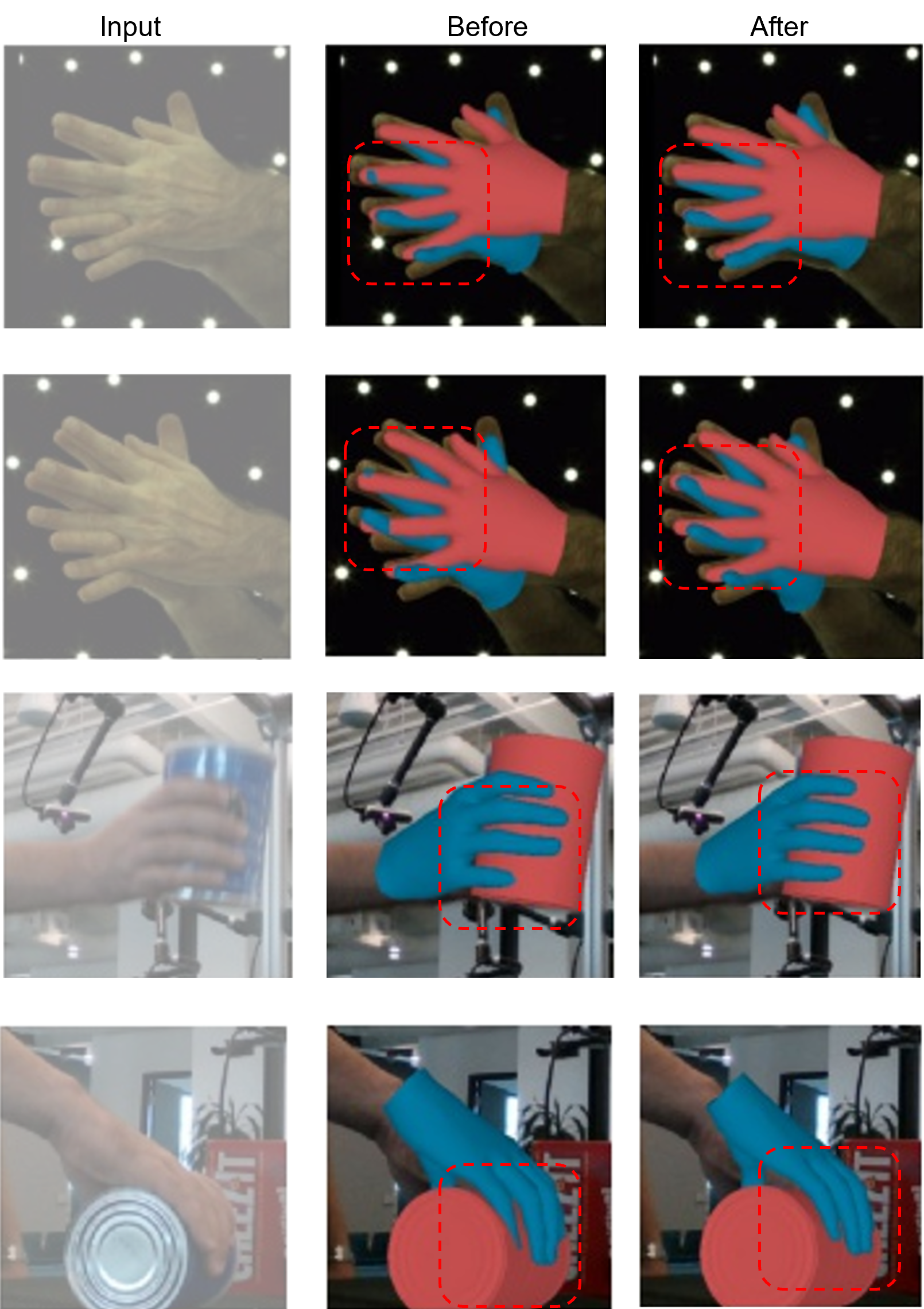}
	\caption{Interaction refinement results. For each triplet, left to right columns correspond to input RGB images, our meshes before and after interaction refinement. Red boxes highlight the interaction refinement regions. }
	\label{fig:contact_region}}
\end{figure}

\subsection{Comparison with the State-of-the-Art}
\paragraph{Quantitative Results.} The comparison with state-of-the-art non-parametric model-based methods~\cite{Li2022intaghand,Kulon_2020_CVPR,lin2021end-to-end} and MANO model based methods~\cite{Zhang2021twohand,hasson19_obman,moon2020interhand2,rong2021monocular,boukhayma20193d,zimmermann2019freihand} is shown in Table~\ref{tab:quantitative_table1}, where the results is based on their released source code and default parameters.  Considering the hand pose and shape accuracy, our integrated model (Ours GCN or Ours Trans) obtains the lowest or second-best MPJPE and MPVPE  on all datasets. Especially compared to the latest MANO model based methods,  our method reduces the pose error  MPJPE by nearly 10\%.  Our MPJPE is comparable to Li et al.~\cite{Li2022intaghand}, despite the attention network used to learn the two-hand features. However, compared to their GCN baseline, our proposed model shows a higher pose and shape accuracy. In addition, regarding the hand mesh plausibility, our proposed model achieves the best holistic performance in terms of edge distance and normal error across all comparisons. Especially compared to the non-parametric model based methods, our method reduces the edge distance error by at least 40\%.  The above quantitative results verify the effectiveness of our integrated model in obtaining well-aligned and plausible hand meshes.

\paragraph{Qualitative Results.} The visualizations of our hand modeling in Fig.~\ref{fig:qualitative1} verify that our proposed model achieves well-aligned and plausible hand reconstructions. Additionally, Fig.~\ref{fig:qualitative2} compares our method to state-of-the-art, demonstrating that our proposed method has lower penetration and yields more plausible two-hand interactions, although ~\cite{Li2022intaghand} achieved better MPJPE and MPVEP than our method by using a complex attention network. Besides, these visualization results also verify that the physical contact loss refinement is better than feature level refinement by using an attention network. More qualitative results are available in the Supplementary.

\paragraph{Interaction Refinement Results.} We also show the quantitative results in Table~\ref{tab:quantitative1} and qualitative results in Fig~\ref{fig:qualitative2}. Our model (Ours Before) achieves the best or second-best performance across all comparisons. In addition, as our proposed model integrates the MANO model, we can leverage the physical contact loss (Ours After) to refine our two-hand or hand-object interactions and reduce M-PD by nearly 50\%. This reveals the effectiveness of our proposed interaction refinement and emphasizes the importance of physical contact loss when compared to the attention feature learning like~\cite{Li2022intaghand} for interaction refinement.

\begin{table}[htb]
\centering
\scalebox{0.85}
{
\begin{tabular}{p{2.7cm}|p{0.7cm}p{0.7cm}p{0.7cm}|p{0.7cm}p{0.7cm}p{0.7cm}} 
\hline
\multicolumn{1}{l|}{Dataset} &\multicolumn{3}{c}{InterHand} &\multicolumn{3}{c}{DexYCB}\\ 
\cline{1-7} 
Method &\scriptsize{MPVPE} &\scriptsize{A-PD} &\scriptsize{M-PD} &\scriptsize{MPVPE} &\scriptsize{A-PD} &\scriptsize{M-PD}
\\\hline
Li et al.~\cite{Li2022intaghand}  &9.03 &1.04 &17.61 &- &- &-\\
MANO-CNN~\cite{zimmermann2019freihand} &14.27 &1.03 &17.81 &11.61 &1.01 &16.78\\ 
GCN-Vert~\cite{Kulon_2020_CVPR}  &10.23 &1.03 &17.95 &9.39 &0.98 &16.95\\ 
Transformer~\cite{lin2021end-to-end}  &10.83&1.04 &18.37 &10.48 &1.05 &17.54\\\hline
GT    &0 &\textbf{0.17} &\textbf{4.89} &0 &\textbf{0.15} &\textbf{3.21} \\ \hline
Ours (Before)  &9.89 &1.00 &17.51 &9.12 &0.94 &16.51\\
Ours (After)  &\textbf{9.92} &\underline{0.51} &\underline{7.62} &\textbf{9.33} &\underline{0.45} &\underline{6.73}\\\hline
\end{tabular}}
\caption{Comparisons between our model (before and after refinement) versus state-of-the-art on InterHand2.6M and DexYCB test sets. \textbf{Best} and \underline{second-best} scores. Our model achieves the best interaction performance across all comparisons. Note that the A-PD and M-PD of InterHand and DexYCB ground truth data are non-zero due to the rigid modeling of both the hand and the object.}
\label{tab:quantitative1}
\end{table}  

\begin{table}[htb]
\centering
\scalebox{1.0}
{
\begin{tabular}{p{2.7cm}|p{0.7cm}p{0.7cm}p{0.7cm} p{0.7cm}} 
\hline
\multicolumn{1}{l|}{Dataset} &\multicolumn{4}{c}{FreiHAND} \\ 
\cline{1-5} 
Method  &\scriptsize{MPJPE} &\scriptsize{MPVPE} &\scriptsize{Edge} &\scriptsize{Norm} \\\hline
Ours(w/o VAE)  &9.13 &9.20 &0.60 &0.18 \\
Ours(w/o Self-dis.)  &8.63 &8.66 &0.58 &0.16 \\
Ours (full)  &\textbf{7.42} &\textbf{7.43} &\textbf{0.51} &\textbf{0.15} \\\hline
\end{tabular}}
\caption{Ablation study on FreiHAND test sets. Best scores are highlighted in \textbf{Bold}.}
\label{tab:ablation}
\end{table} 

\subsection{Ablation  Studies}
\textbf{VAE Module.} We also compare among our baseline models (Ours w/o VAE) in Table~\ref{tab:ablation}. Our model (w/o VAE) is a pipeline without refining the joints, which directly use a linear regress matrix to covert non-parametric model meshes to the MANO model joints space. Our full model outperforms this baseline by nearly 20\%, which verifies the effectiveness of the VAE refinement module.

\textbf{Self-distillation learning.} The impact of self-distillation learning is shown in Table~\ref{tab:ablation} (Ours w/o Self-dis.). Self-distillation learning reduces pose and shape error by nearly 10\%. Furthermore, our full model reduces the pose and shape errors by nearly 50\% compared to the non-parametric model based method, \ie, GCN-joint in Table~\ref{tab:quantitative_table1}, which only uses joint as supervision. This reveals the effectiveness of our self-distillation learning and integrated strategy.

\textbf{Analysis of the twist rotation.}\label{sec:twist}
To evaluate the effectiveness of the twist rotation, besides our estimated twist from the network (Estimated Twist), we also set the twist as zero (Zero Twist) or a random value from 0 to 1 (Random Twist).
The results are given in Table~\ref{tab:twist}. Firstly, the performance of the Zero Twist is comparable to that of our Estimated Twist. These results are reasonable since most hand joints' twist rotation angles are close to zero. In contrast, there is a considerable performance gap between the Random Twist results and our Estimated Twist results, which shows the necessity of twist rotation estimation.
\begin{table}[htb]
\centering
\scalebox{1.0}
{
\begin{tabular}{p{2.7cm}|p{0.7cm}p{0.7cm}|p{0.7cm} p{0.7cm}} 
\hline
\multicolumn{1}{l|}{Dataset} &\multicolumn{2}{c}{FreiHand}&\multicolumn{2}{c}{DexYCB} \\
\cline{1-5} 
Method  &\scriptsize{MPJPE} &\scriptsize{MPVPE} &\scriptsize{MPJPE} &\scriptsize{MPJPE} \\\hline
Random Twist  &12.68 &13.90 &11.78 &12.52 \\
Zero Twist  &7.83 &7.96 &9.45 &9.67 \\
Estimated Twist &\textbf{7.42} &\textbf{7.43} &\textbf{8.92} &\textbf{9.12} \\\hline
\end{tabular}}
\caption{Reconstruction error with different twist angles.  Best scores are highlighted in \textbf{Bold}.}
\label{tab:twist}
\end{table} 

 \subsection{Limitations}
\begin{figure}[!htb]
	\centering{
	\includegraphics[width=1.0\linewidth]{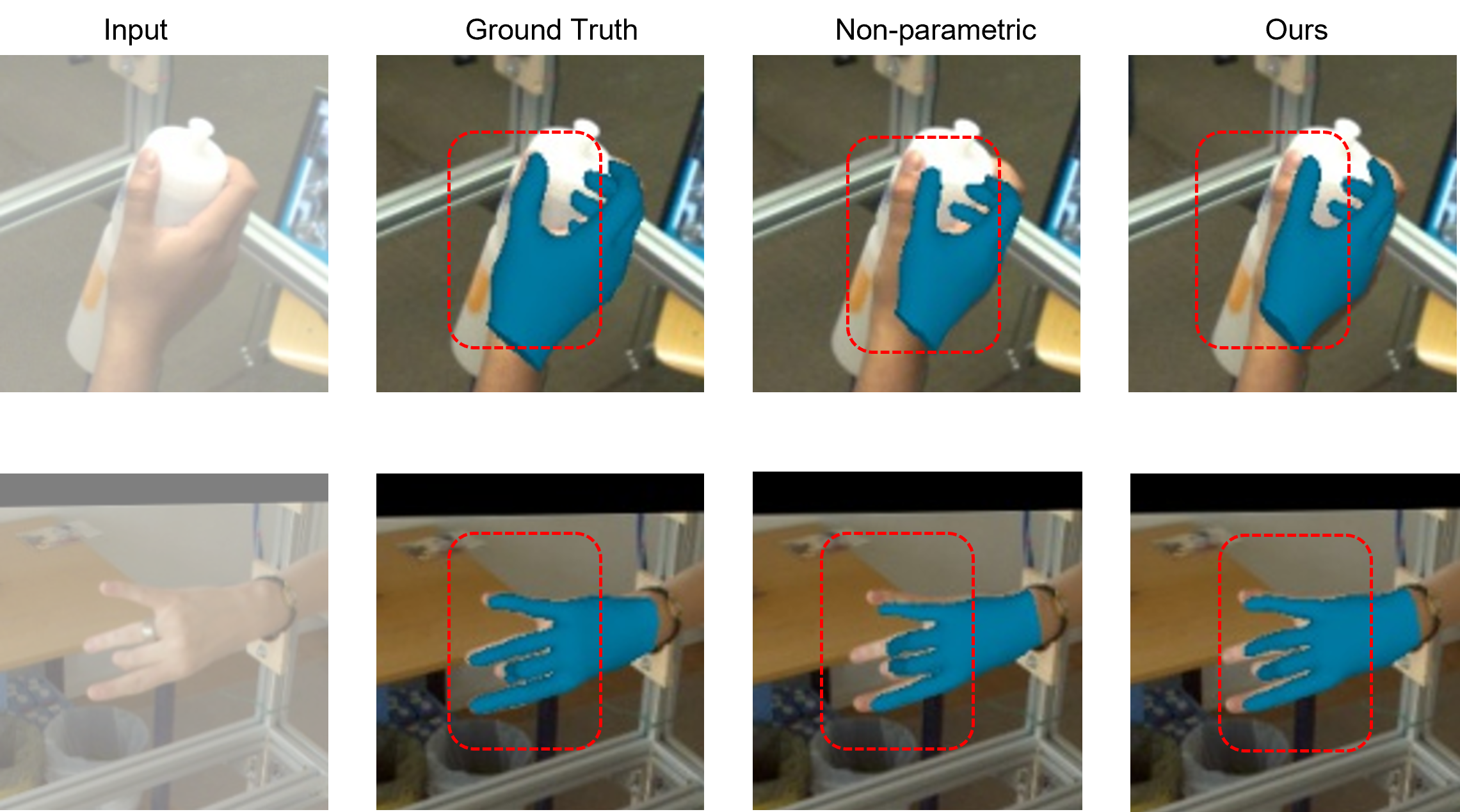}
	\caption{Limitation results. For each row, the left to right columns correspond to input RGB, ground truth, our non-parametric model mesh and our final mesh. We are limited by our initial hand mesh from our non-parametric model. Red boxes highlight the not aligned regions.}
	\label{fig:limitations}}
\end{figure}

Our pose decoder pipeline uses a non-parametric model to obtain the initial meshes. These initial hand meshes limit our results compared to the ground truth (see Fig.~\ref{fig:limitations}). Although our initial hand meshes are not well aligned, our output meshes are close to the ground truth and better than our initial hand meshes, which verifies the effectiveness of our integrated model on the other side.  
These limitations can be improved by considering other sources of information, like rendered masks, to offer extra supervision to improve the initial hand mesh accuracy.

\section{Conclusion}
This work proposes an effective integrated framework of a non-parametric model and MANO model for estimating well-aligned and plausible hand meshes from RGB images. We explore the trade-off between the non-parametric and MANO model for hand surface modelling and propose the first integrated model to overcome this trade-off. Additionally, to improve the accuracy of hand meshes and mitigate the gap between the non-parametric model joints and MANO deformation joints, we introduce a VAE to solve it. Furthermore, we introduce a self-distillation learning method that utilizes our parametric mesh to boost the non-parametric model's mesh learning. 
Experimental results show that our proposed method achieves better performance over existing MANO-based and non-parametric model based hand shape estimation methods, on single-hand task, two-hand interaction or hand-object interaction task. This verifies the effectiveness of our integrated framework of a non-parametric model and a parametric model. In future work, we will explore using a render mask as extra supervision to improve the hand shape modeling based on our integrated model.

\paragraph{Acknowledgements}
This research is supported by the National Research Foundation, Singapore and DSO National Laboratories under its AI Singapore Programme (AISG Award No: AISG2-RP-2020-016). Any opinions, findings and conclusions or recommendations expressed in this material are those of the author(s) and do not reflect the views of National Research Foundation, Singapore. 
{\small
\bibliographystyle{ieee_fullname}
\bibliography{cvprarxiv}
}

\end{document}